\documentclass{article}

    \PassOptionsToPackage{numbers, compress}{natbib}

    \usepackage[preprint]{neurips_2024}

\usepackage[utf8]{inputenc} %
\usepackage[T1]{fontenc}    %
\usepackage{hyperref}       %
\usepackage{url}            %
\usepackage{booktabs}       %
\usepackage{amsfonts}       %
\usepackage{nicefrac}       %
\usepackage{microtype}      %
\usepackage[table]{xcolor}
\definecolor{lightpink}{rgb}{1.0, 0.71, 0.76}
\definecolor{lightblue}{rgb}{0.68, 0.85, 0.9}

\usepackage{graphicx}
\usepackage{caption}
\usepackage{multirow}
\usepackage{amsmath}  %
\usepackage{amsfonts} %
\usepackage{amssymb}  %
\usepackage{bm}       %
\usepackage{makecell}
\usepackage{hyperref}

\title{$\textit{S}^3$Gaussian: Self-Supervised Street Gaussians for Autonomous Driving}

\DeclareUrlCommand\url{\color{magenta}}

\author{
Nan Huang$^{1,2,}$\footnotemark[1] \quad
Xiaobao Wei$^{2}$ \quad
Wenzhao Zheng$^{1,3,}$\footnotemark[2]  \quad 
Pengju An$^2$\quad 
Ming Lu$^{2}$ \\ \quad 
\textbf{Wei Zhan}$^{1}$\quad 
\textbf{Masayoshi Tomizuka}$^{1}$\quad 
\textbf{Kurt Keutzer}$^{1}$\quad 
\textbf{Shanghang Zhang}$^{2,}$\footnotemark[3] \\
\url{https://wzzheng.net/S3Gaussian}\\
$^1$UC Berkeley \quad
$^2$Peking University \quad 
$^3$Tsinghua University \\
\texttt{wenzhao.zheng@outlook.com; shanghang@pku.edu.cn}
}

\begin{document}

\maketitle

\renewcommand{\thefootnote}{\fnsymbol{footnote}}
\footnotetext[1]{Work done during an internship at UC Berkeley. $\dagger$Project leader. $\ddagger$Corresponding author.}
\renewcommand{\thefootnote}{\arabic{footnote}}

\begin{abstract}
Photorealistic 3D reconstruction of street scenes is a critical technique for developing real-world simulators for autonomous driving.
Despite the efficacy of Neural Radiance Fields (NeRF) for driving scenes, 3D Gaussian Splatting (3DGS) emerges as a promising direction due to its faster speed and more explicit representation. 
However, most existing street 3DGS methods require tracked 3D vehicle bounding boxes to decompose the static and dynamic elements for effective reconstruction, limiting their applications for in-the-wild scenarios.
To facilitate efficient 3D scene reconstruction without costly annotations, we propose a self-supervised street Gaussian ($\textit{S}^3$Gaussian) method to decompose dynamic and static elements from 4D consistency.
We represent each scene with 3D Gaussians to preserve the explicitness and further accompany them with a spatial-temporal field network to compactly model the 4D dynamics.
We conduct extensive experiments on the challenging Waymo-Open dataset to evaluate the effectiveness of our method.
Our $\textit{S}^3$Gaussian demonstrates the ability to decompose static and dynamic scenes and achieves the best performance without using 3D annotations. 
Code is available at: \url{https://github.com/nnanhuang/S3Gaussian/}.

\end{abstract}

\section{Introduction}
Autonomous driving has made significant progress in recent years and developed various techniques in each stage of its pipeline including perception~\cite{bevformer, beverse, tpvformer, surroundocc}, prediction~\cite{hu2021fiery,gu2022vip3d,liang2020pnpnet}, and planning~\cite{Dauner2023CORL,cheng2022gpir,cheng2023forecast}.
With the emergence of end-to-end autonomous driving which directly outputs the control signal from sensor inputs~\cite{hu2022stp3,hu2023uniAD,jiang2023vad}, open-loop evaluation of autonomous driving systems ceases to be effective and thus requires pressing improvement~\cite{zhai2023rethinking,li2023ego}.
As a promising solution, real-world closed-loop evaluation requires sensor inputs for controllable views, which motivates the development of high-quality scene reconstruction methods~\cite{turki2023suds, xie2023s}.

Despite numerous efforts on photo-realistic reconstruction on small-scale scenes~\cite{mildenhall2021nerf, muller2022instant, chen2022tensorf, kerbl20233d, wei2023noc}, the large-scale and highly dynamic characteristics of driving scenarios pose new challenges to the effective modeling of 3D scenes.
To accommodate these, most existing works adopt tracked 3D bounding boxes to decompose static and dynamic elements~\cite{Yan2024StreetGF, wu2023mars, turki2023suds}.
Still, the costly annotations of 3D tracklets limit their applications for 3D modeling from in-the-wild data.
EmerNerf~\cite{yang2023emernerf} addressed this by simultaneously learning the scene flow and using it to connect corresponding points in the 4D NeRF field for multi-frame reconstruction, enabling the emergence of decomposition between static and dynamic objects without explicit bounding boxes.
However, 3D driving scene modeling has been undergoing a shift from NeRF-based reconstruction to 3D Gaussian Splatting due to its desire for low latency and explicit representation.
Though EmerNerf demonstrated promising results, it can only be used for NeRF-based scene modeling, which takes a long time for training and rendering. 
It is still unclear how to achieve 3D Gaussian Splatting for urban scene reconstruction without explicit 3D supervision.

To address the above issues, we propose a \textbf{S}elf-\textbf{S}upervised \textbf{S}treet \textbf{Gaussians} named $S^3$Gaussian, offering a robust solution for dynamic street scenes without requiring 3D supervision. Specifically, to handle the complex spatial-temporal deformations inherent in driving scenes, $S^3$Gaussian introduces a cutting-edge spatial-temporal field for scene decomposition in a self-supervised manner. This spatial-temporal field incorporates a multi-resolution Hexplane structure encoder alongside a compact multi-head Gaussian decoder. The Hexplane encoder is designed to decompose the 4D input grid into multi-resolution, learnable feature planes, efficiently aggregating temporal and spatial information from the dynamic street scenes. During the optimization process, the multi-resolution Hexplane structure encoder effectively separates the entire scene, achieving a canonical representation for each scene. Dynamic-related features are stored within the spatial-temporal plane, while static-related features are retained in the spatial-only plane. Leveraging the densely encoded features, the multi-head Gaussian decoders calculate the deformation offsets from the canonical representations. These deformations are then added to the original 3D Gaussians' attributes, including position and spherical harmonics, allowing for a dynamic alteration of the scene representation conditioned on time series. Our main contributions are summarized as follows:

\begin{itemize}

\item We propose $S^3$Gaussian, the first self-supervised method that manages to decompose the dynamic and static 3D Gaussians in street scenes without extra manually annotated data.

\item To model the complex changes in driving scenes, we introduce an efficient spatial-temporal decomposition network to automatically capture the deformation of 3D Gaussians.

\item We conduct comprehensive experiments on challenging datasets, including NOTR and Waymo. Results demonstrate that $S^3$Gaussian achieves state-of-the-art rendering quality on scene reconstruction and novel view synthesis tasks.

\end{itemize}

\begin{figure*}
    \centering
    \includegraphics[width=1\textwidth]{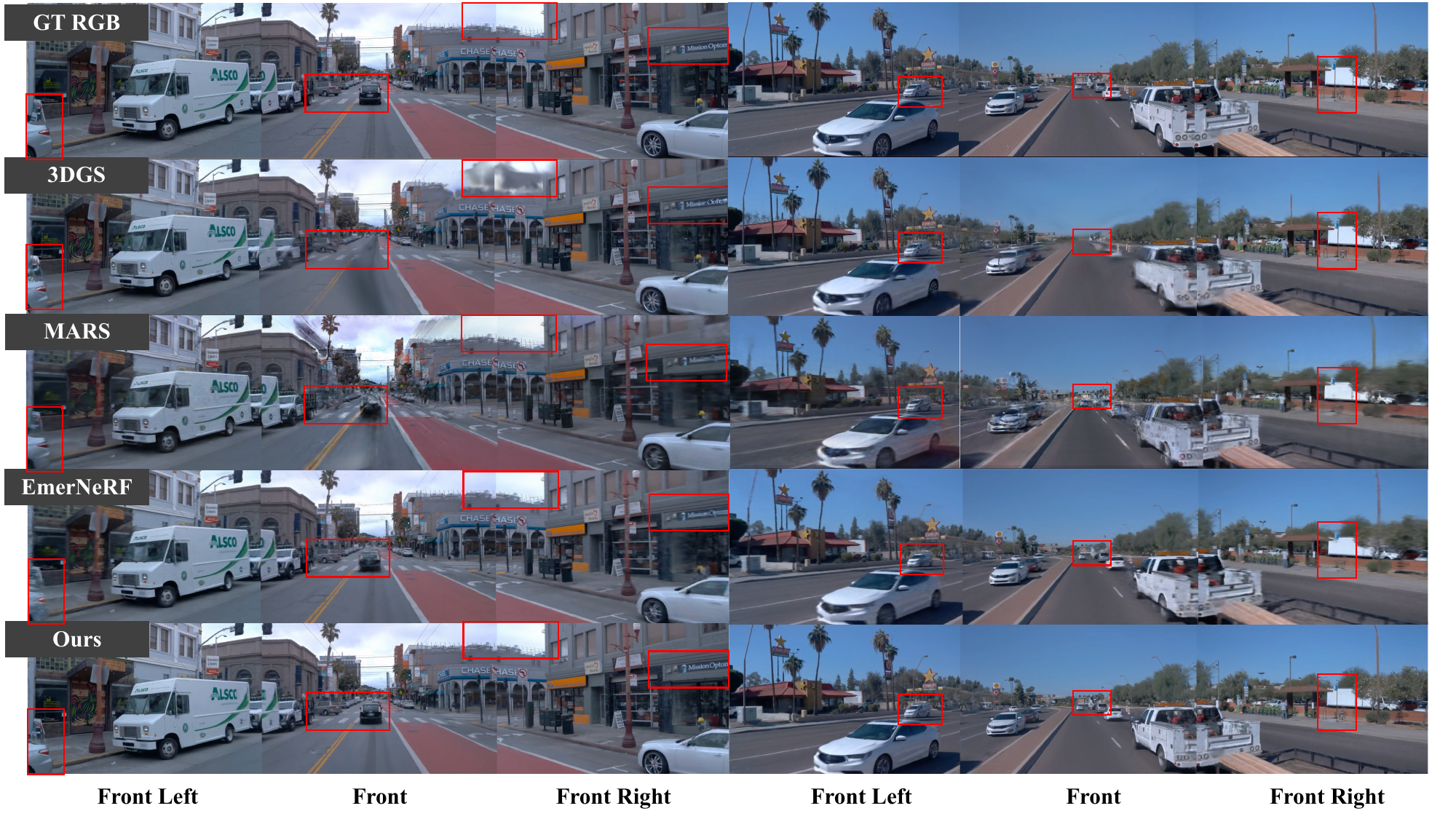}
    \vspace{-2em}
    \caption{Qualitative comparison over Waymo-NOTR Datasets. On the left, we showcase results from novel view synthesis; on the right, results from dynamic scene reconstruction are displayed. With the proposed spatial-temporal network for the self-supervised scene decomposition, our method $S^3$Gaussian produces the best rendering quality with high fidelity and sharp details.}
    \label{fig:compare}
    \vspace{-2em}
\end{figure*}

\vspace{-1em}

\section{Related Work}

\textbf{3D Gaussian Splatting.}
Recent breakthroughs in 3D Gaussian Splatting (3DGS)~\cite{kerbl3Dgaussians} have revolutionized scene modeling and rendering. Harnessing the power of explicit 3D Gaussians, 3DGS achieves optimal outcomes in novel view synthesis and real-time rendering while also substantially reducing parameter complexity compared to conventional representations such as meshes or voxels. This technique seamlessly integrates the principles of point-based rendering~\cite{Aliev2019NeuralPG} and splatting~\cite{Zwicker2002EWAS}, facilitating rapid rendering and differentiable computation through splat-based rasterization.

While the original 3DGS model is designed for static scene representation, several researchers have extended its applicability to dynamic objects and scenes. For instance, Yang et al.~\cite{Yang2023Deformable3G} introduces a deformation network aimed at capturing Gaussian motion from a series of dynamic monocular images. Another approach, detailed by~\cite{Wu20234DGS}, establishes connections between neighboring Gaussians using a HexPlane, thereby enabling real-time rendering. By optimizing point clouds containing semantic logits and 3D Gaussians for novel dynamic scene representation, Yan et al.~\cite{Yan2024StreetGF} achieves improvements in training and rendering speed. Similarly to NeRF's methodology, Zhou et al.~\cite{Zhou2023DrivingGaussianCG} differentiates static backgrounds and dynamic objects within the scene and reconstructs each using distinct Gaussian Splatting methods. However, existing approaches are constrained as they can model only static or dynamic scenes individually and require supervised classification of scene types. Our objective is to autonomously learn the decomposition of static and dynamic scenes in a self-supervised manner, thereby eliminating the reliance on real annotations, such as dynamic object bounding boxes.

\textbf{Street Scene Reconstruction for Autonomous Driving Simulation.}
Numerous efforts have been put into reconstructing scenes from autonomous driving data captured in real scenes. 
Existing self-driving simulation engines such as CARLA~\cite{Dosovitskiy2017CARLAAO} or AirSim~\cite{Shah2017AirSimHV} suffer from costly manual effort to create virtual environments and the lack of realism in the generated data. 
The rapid development of Novel View Synthesis (NVS) techniques, including NeRF~\cite{mildenhall2021nerf} and 3DGS ~\cite{kerbl3Dgaussians}, has attracted considerable attention within the arena of autonomous driving. 
Many studies~\cite{Chen2023PeriodicVG, Guo2023StreetSurfEM, Liu2023RealTimeNR, Lu2023UrbanRF, Ost_Mannan_Thuerey_Knodt_Heide_2021, Rudnev2021NeRFFO, Rematas2021UrbanRF, Tancik2022BlockNeRFSL, Tonderski2023NeuRADNR, turki2023suds, Turki2021MegaNeRFSC, wu2023mars, Yan2024StreetGF, Yang2023UniSimAN, Zhou2023DrivingGaussianCG} have investigated the application of these methods for reconstructing street scenes. 
Block-NeRF~\cite{Tancik2022BlockNeRFSL} and Mega-NeRF~\cite{Turki2021MegaNeRFSC} propose segmenting scenes into distinct blocks for individual modeling. Urban Radiance Field~\cite{Rematas2021UrbanRF} enhances NeRF training with geometric information from LiDAR, while DNMP~\cite{Lu2023UrbanRF} utilizes a pre-trained deformable mesh primitive to represent the scene. Streetsurf~\cite{Guo2023StreetSurfEM} divides scenes into close-range, distant-view, and sky categories, yielding superior reconstruction results for urban street surfaces. For modeling dynamic urban scenes, NSG~\cite{Ost2020NeuralSG} represents scenes as neural graphs, and MARS~\cite{wu2023mars} employs separate networks for modeling background and vehicles, establishing an instance-aware simulation framework. With the introduction of 3DGS~\cite{kerbl3Dgaussians}, DrivingGaussian~\cite{Zhou2023DrivingGaussianCG} introduces Composite Dynamic Gaussian Graphs and incremental static Gaussians, while StreetGaussian~\cite{Yan2024StreetGF} optimizes the tracked pose of dynamic Gaussians and introduces 4D spherical harmonics for varying vehicle appearances across frames. 

The aforementioned methods not only suffer from prolonged training durations and sluggish rendering speeds but also fail to qualify the ability to divide dynamic and static scenes automatically.
Therefore, we propose $S^3$Gaussian to differentiate between dynamic and static scenes in a self-supervised manner without the need for additional annotations, and perform high-fidelity and real-time neural rendering of dynamic urban street scenes, which is crucial for autonomous driving simulation.

\section{Proposed Approach}
\label{sec/methods}
We aim to learn a spatial-temporal representation of the dynamic environment of the street from a sequence of images captured by moving vehicles. However, due to the limited number of observation views and the high cost of obtaining ground truth annotations for dynamic and static objects, we aim to learn the scene decomposition of both static and dynamic components in a fully self-supervised manner, avoiding the supervision of extra annotations including bounding boxes for dynamic objects, segmentation masks for the scene decomposition, and optical flow for the motion perception.

To achieve these objectives, we propose a novel scene representation named $\textit{S}^3$Gaussian. First, in Sec.~\ref{sec/4d gaussians}, we lift 3D Gaussians to 4D to better represent dynamic and complex scenes. Then, in Sec.~\ref{sec/field network}, we introduce a novel Spatial-temporal Field Network to integrate high-dimensional spatial-temporal information and decode them to transform 4D Gaussians. Finally, in Sec.~\ref{sec/optimize}, we describe the entire optimization process, eliminating extra annotations.

\begin{figure*}[t]
    \centering
    \includegraphics[width=1\textwidth]{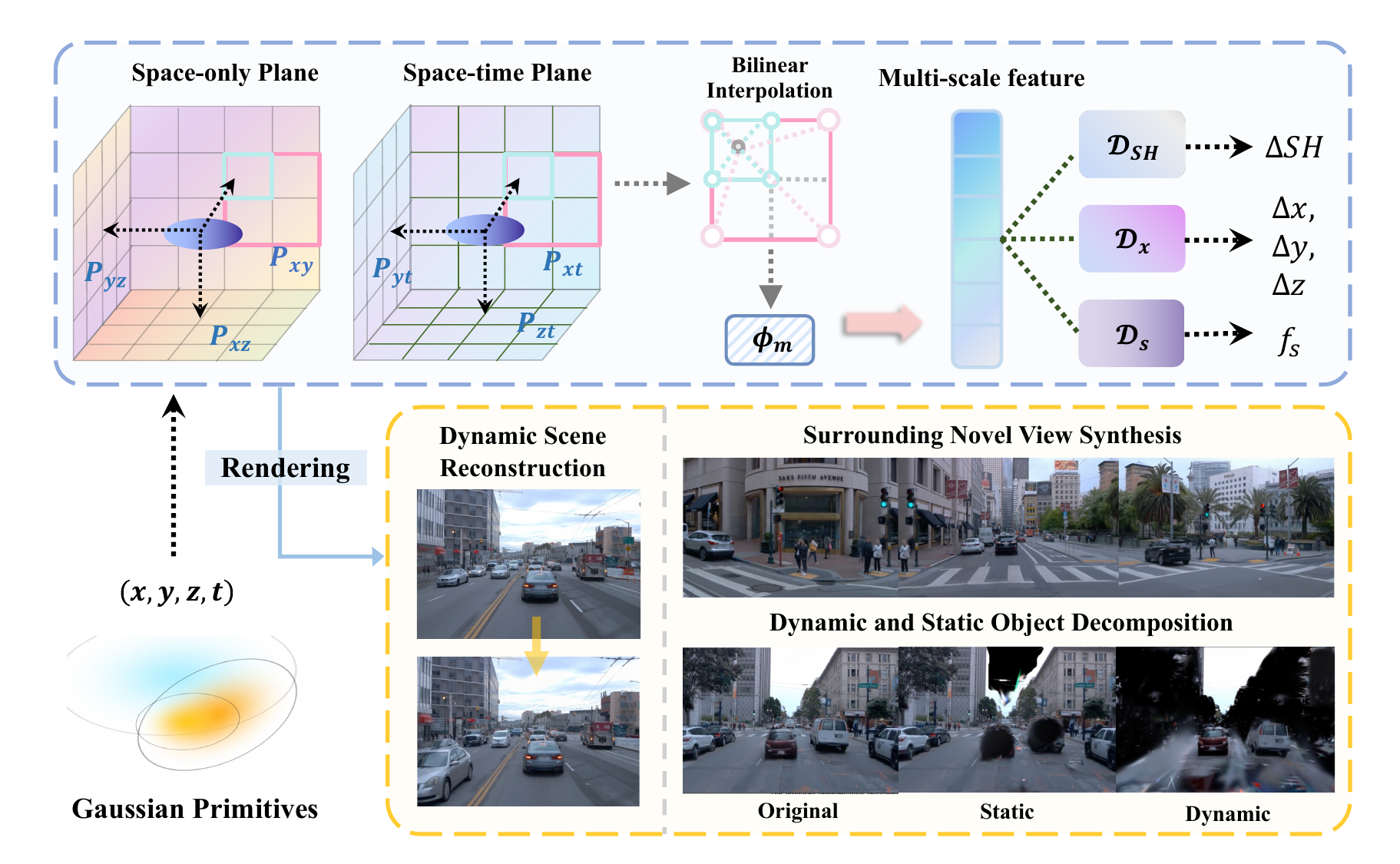}
    \vspace{-2em}
    \caption{Pipeline of $S^3$Gaussian. To tackle the challenges in self-supervised street scene decomposition, our method consists of a Multi-resolution Hexplane Structure Encoder to encode 4D grid into feature planes and a multi-head Gaussian Decoder to decode them into deformed 4D Gaussians. The entire pipeline is optimized without extra annotations in a self-supervised manner, leading to superior scene decomposition ability and rendering quality.}
    \label{fig:pipeline}
    \vspace{-1.5em}
\end{figure*}

\subsection{4D Gaussian Representations}
\label{sec/4d gaussians}
As depicted in Figure~\ref{fig:pipeline}, our scene representations include 3D Gaussians~\cite{kerbl3Dgaussians} $\mathcal{G}$ and a Spatial-temporal Field Network $\mathcal{F}$.
To depict static scenes, 3D Gaussians are characterized by a covariance matrix $\Sigma$ and a position vector $\mathcal{X}$, referred to as the geometric attributes. For a stable optimization, each covariance matrix is further factorized into a scaling matrix $\mathcal{S}$ and a rotation matrix $\mathcal{R}$:
\vspace{-1mm}
\begin{equation}
\Sigma = \mathcal{R} \mathcal{S} \mathcal{S}^{T} \mathcal{R}^{T}
\label{eq:cov matrics}
\end{equation}
In addition to the position and covariance matrices, each Gaussian is also assigned an opacity value $\alpha \in \mathbb{R}$ and color $\mathcal{C} \in \mathbb{R}^{3(k+1)^2}$, defined by spherical harmonic (SH) coefficients, where $k$ represents the degrees of SH functions.

The Spatial-temporal Field Network takes the position of each Gaussian $\mathcal{X}$ and the current timestep $t$ as input, producing spatial-temporal features $f$. After decoding these features, the network can predict the displacement $\bigtriangleup \mathcal{G}$ of each point relative to canonical space while also obtaining semantic information $f_s$ through the semantic feature decoder $\mathcal{D}_s$. We detail it in Sec.~\ref{sec/field network}.

Following~\cite{Yifan_2019}, we utilize a differentiable 3D Gaussian splatting renderer $\mathcal{R}$ to project the deformed 3D Gaussians $\mathcal{G}' = \bigtriangleup \mathcal{G} + \mathcal{G} $ into 2D~\cite{zwicker2001surface}. Here, the covariance matrix $\Sigma'$ in camera coordinates is:
\vspace{-1mm}
\begin{equation}
\Sigma ' = JW\Sigma W^TJ^T
\label{eq:2d cov matrics}
\end{equation}
where $J$ is the Jacobian matrix of the perspective projection, and $W$ is the viewing transform matrix. The color of each pixel is calculated by $N$ ordered points using $\alpha$-blending:
\begin{equation}
C = \sum_{i \in N}^{} c_i\alpha _i\prod_{j=1}^{i-1} (1-\alpha _i)
\label{eq:color}
\end{equation}
Here, $\alpha_i$ and $c_i$ represent the opacity and color of one point, computed by an optimizable per-point opacity and SH color coefficients with the view direction.
The semantic map can be rendered simply by changing the color $c$ in Eq.~\ref{eq:color} to the semantic feature $f_s$.

\subsection{Spatial-temporal Field Network}
\label{sec/field network}
The primary focus of vanilla 3D Gaussians Splatting is on tasks in static scenes. However, the real world is dynamic, especially in contexts like autonomous driving. This makes the transition from 3DGS to 4D a crucial and challenging endeavor. Firstly, in dynamic scenarios, the views captured by each moving camera at each time step are sparser than in static scenes, making individual modeling of each time step exceptionally difficult due to this sparsity. Therefore, it becomes imperative to consider information sharing across time steps~\cite{Fridovich-Keil_Meanti_Warburg_Recht_Kanazawa_2023}.

Moreover, modeling all Gaussian points in space and time is impractical for large-scale or long-duration scenarios like autonomous driving due to significant memory overhead. Hence, we propose leveraging an efficient Gaussian-based spatial-temporal network to model 3D Gaussian motion. This network comprises a Multi-resolution Hexplane Structure Encoder and a minimal Multi-head Gaussian Decoder. It only needs to maintain a set of canonical 3D Gaussians and model a deformation field for each timestep. This field predicts displacement and color changes relative to the canonical space 3D Gaussians, thus capturing Gaussian motion~\cite{Wu20234DGS}. Additionally, we incorporate a simple semantic field to assist in automatically decomposing static and dynamic Gaussians.

\textbf{Multi-resolution Hexplane Structure Encoder.}
To efficiently aggregate temporal and spatial information across timesteps, considering that adjacent Gaussians often share similar spatial and temporal characteristics, we employ the Multi-resolution Hexplane Structure Encoder $\mathcal{E} $ with a tiny MLP $\phi _m$ to represent dynamic 3D scenes effectively inspired by~\cite{Cao_Johnson,Fang_Nießner,Fridovich-Keil_Meanti_Warburg_Recht_Kanazawa_2023,Shao_Zheng_Tu_Liu_Zhang_Liu_2022}. Specifically, the HexPlane decomposes the 4D spatial-temporal grid into six multi-resolution learnable feature planes spanning each pair of coordinate axes, each endowed with an orthogonal axis. The first three planes $\mathcal{P}  _{xy}$, $\mathcal{P}  _{xz}$, $\mathcal{P}  _{yz}$ represent spatial-only dimensions, while the latter three $\mathcal{P}  _{xt}$, $\mathcal{P}  _{yt}$, $\mathcal{P}  _{zt}$ represent spatial-temporal variations. This decoupling of time and space is beneficial for separating static and dynamic elements. Dynamic objects become distinctly visible on the spatial-temporal plane, while static objects solely manifest on the spatial-only plane.

Additionally, to promote spatial smoothness and coherence while compressing the model and reducing the number of features stored at the highest resolution, inspired by Instant-NGP's multi-scale hash encoding~\cite{Müller_Evans_Schied_Keller_2022}, our hexplane encoder comprises multiple copies of different resolutions. This representation effectively encodes spatial features at various scales. Therefore, our formulation is:
\vspace{-1.5mm}
\begin{equation}
\mathcal{P}^{\rho}_{ij} \in \mathbb{R} ^{d\times \rho r_i \times \rho r_j}, (i,j) \in 
 \{   (x,y), (x,z), (y,z), (x,t), (y,t), (z,t) \}
 ,\rho \in \{ 1,2 \}
\label{eq:hexplane}
\end{equation}
where $d$ is the hidden dimension of features, $\rho$ stands for the upsampling scale, and $r$ equals to the basic resolution. Giving a 4D coordinate $(x,y,z,t)$, we then obtain the neural voxel features and merge all the features using a tiny MLP $\phi _m$ as follows:
\vspace{-1.5mm}
\begin{equation}
f(x,y,z,t) =\phi _m( \bigcup_{\rho}^{}  \prod_{}^{} \pi  (\mathcal{P}^{\rho}_{ij} , \psi^{\rho}_{ij}(x,y,z,t) ))
\label{eq:feature}
\end{equation}
where $\psi^{\rho}_{ij}$ projects 4D coordinate $(x,y,z,t)$ onto the corresponding plane, and $\pi$ denotes bilinear interpolation, used for querying voxel features located at the four vertices.
We merge the planes using Hadamard product to produce spatially localized signals, as discussed in~\cite{Fridovich-Keil_Meanti_Warburg_Recht_Kanazawa_2023}.

\textbf{Multi-head Gaussian Decoder.}
We use separate MLP heads $\mathcal{D} =(\mathcal{D} _{SH},\mathcal{D}_{x} , \mathcal{D}_{s} )$ to decode the features obtained in Sec.~\ref{sec/field network}. Specifically, we employ a semantic feature decoder to compute semantic features $f_s = \mathcal{D}_s (f(x,y,z,t))$. Considering that most autonomous driving scenarios involve rigid motion, we only consider deformation in the position of the Gaussians, thus $\triangle x = \mathcal{D}_x (f(x,y,z,t))$. Additionally, considering factors like illumination, the appearance of the scene varies with its global position and time. Therefore, we also introduce an SH coefficient head to model the 4D dynamic appearance model $\triangle SH = \mathcal{D}_{SH} (f(x,y,z,t))$. Finally, our deformed 4D Gaussians are formulated as: $\mathcal{G}' = \{ \mathcal{X} + \triangle \mathcal{X},  \mathcal{C} + \triangle \mathcal{C} , s,r,\sigma ,f_s\} $.

\subsection{Self-supervised Optimization}
\label{sec/optimize}
\textbf{LiDAR Prior Initialization.}
To initialize the positions of the 3D Gaussians, we leverage the LiDAR point cloud captured by the vehicle instead of using the original SFM~\cite{Schonberger_Frahm_2016} point cloud to provide a better geometric structure. To reduce model size, we also downsample the entire point cloud by voxelizing it and filtering out points outside the image. For colors, we initialize them randomly.

\textbf{Optimization Objective.}
The loss function of our method consists of seven parts, and we jointly optimize our scene representation and Spatial-temporal field using it. $\mathcal{L}_{rgb}$ is the L1 loss between rendered and ground truth images and $\mathcal{L}_{ssim}$ measures the similarity between them. $\mathcal{L}_{depth}$ is the L2 loss between the estimated depth map from the LiDAR point cloud and the rendered depth map, used to supervise the expected position of the Gaussians~\cite{yang2023emernerf, Zhou2023DrivingGaussianCG}. The rendered depth is computed using the positions of the Gaussians. $\mathcal{L}_{feat}$ is the L2 loss of semantic feature. Following~\cite{Fang_Nießner,Sun_Sun_Chen_2022,Fridovich-Keil_Meanti_Warburg_Recht_Kanazawa_2023}, we also introduce a grid-based total-variational loss $\mathcal{L}_{tv}$. 
Given that most elements in the scene are static, we introduce regularization constraints into the spatial-temporal network to enhance the separation of static and dynamic components. We achieve this by minimizing the expectation of $\mathbb{E} (\triangle \mathcal{X})$ and $\mathbb{E} (\triangle \mathcal{C})$, which encourages the network only to produce offset values when necessary. Then, the total loss function can be formulated as follows:
\begin{equation}
\mathcal{L} = \lambda_{rgb}\mathcal{L}_{rgb} + \lambda_{depth}\mathcal{L}_{depth}  +  \lambda_{feat}\mathcal{L}_{feat}+ \lambda_{ssim}\mathcal{L}_{ssim} + \lambda_{tv}\mathcal{L}_{tv}+  \lambda_{reg}^x \mathcal{L}_{reg}^x +\lambda_{reg}^y \mathcal{L}_{reg}^{c} 
\label{eq:loss}
\end{equation}
where $\lambda_{rgb}=1.0$, $\lambda_{depth}=0.1$, $\lambda_{feat}=0.1$, $\lambda_{ssim}=0.1$, $\lambda_{tv}=0.1$, $\lambda_{reg}^x=0.01$, and $\lambda_{reg}^y=0.01$ are the weights assigned to each loss component.

\section{Experiments}
\label{exp}
In this section, we primarily discuss the experimental methodology used to evaluate the performance of our $\textit{S}^3$Gaussian. Details of the dataset settings, baseline methods, and implementation specifics are provided in Sec.~\ref{exp:set up}. In Sec.~\ref{exp: sota}, we compare our approach with state-of-the-art (SOTA) methods across various tasks. Further ablation studies and analysis are detailed in Sec.~\ref{exp: ablation}. 

\subsection{Experimental Setup}
\label{exp:set up}

\begin{figure*}[t]
    \centering
    \vspace{-1em}
    \includegraphics[width=1\textwidth]{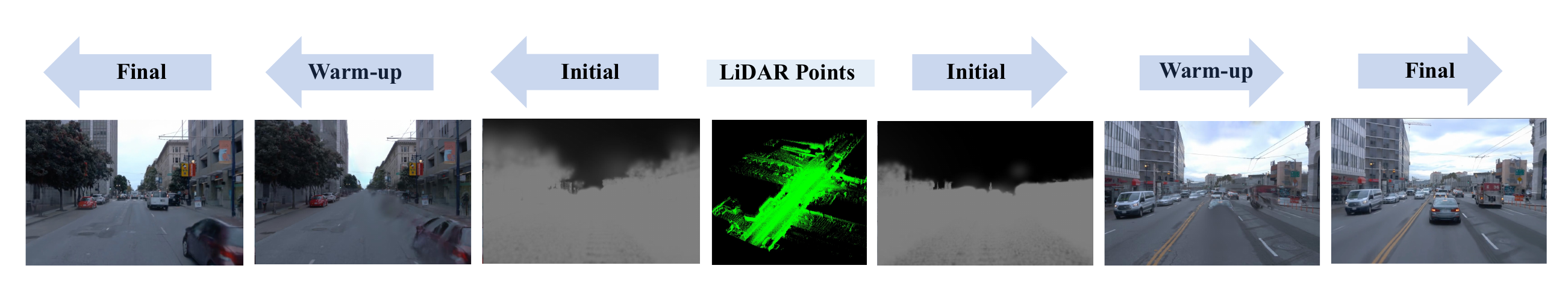}
    \vspace{-2em}
    \caption{Illustration of the optimization process. With the LiDAR points initialization and the static 3D Gaussian Warm-up strategy, our model achieves high-quality 4D Gaussian representations of the complex dynamic scenes.}
    \label{fig:warmup}
    \vspace{-1.5em}
\end{figure*}

\textbf{Datasets.} 
NOTR dataset is a subset of the Waymo Open dataset~\cite{sun2020scalability} curated by~\cite{yang2023emernerf}, which comprises many challenging driving scenarios: ego-static, high-speed, exposure mismatch, dusk/dawn, gloomy, rainy, and night scenes. In contrast, many public datasets with LiDAR data suffer from a severe imbalance, eg. nuScenes~\cite{Caesar_Bankiti_Lang_Vora_Liong_Xu_Krishnan_Pan_Baldan_Beijbom_2020} and nuPlan~\cite{Caesar_Kabzan_Tan_Kit_Wolff_Lang_Fletcher_Beijbom_Omari_2021}, predominantly featuring simple scenes with few dynamic objects. Therefore, we utilize NOTR's dynamic32 (D32) and static32 (S32) datasets, totaling 64 scenes, to obtain a balanced and diverse standard for evaluating our static and dynamic reconstruction.
Furthermore, since most baseline methods are NeRF-based, to ensure a fair evaluation of our method's performance, we conduct comparisons with the current state-of-the-art Gaussian-based method, StreetGaussian~\cite{Yan2024StreetGF}. We adhere to the dataset configuration used by StreetGaussian, employing the six scenes selected from the Waymo Open dataset~\cite{sun2020scalability}, which are characterized by complex environments and significant object motion.

\textbf{Baseline Methods.}
We evaluate our approach against state-of-the-art methods, including NeRF-based models and 3DGS-based models. MARS~\cite{wu2023mars} is a modular~\cite{Tancik_2023} simulator based on NeRF, utilizing 2D bounding boxes to train NeRF for static and dynamic objects respectively. NSG~\cite{Ost_Mannan_Thuerey_Knodt_Heide_2021} learns latent codes to model moving objects with a shared decoder. EmerNeRF~\cite{yang2023emernerf} also builds upon NeRF but self-supervises the modeling of dynamic scenes by optimizing flow fields, representing the current SOTA in self-supervised learning for dynamic driving scene representations. The 3DGS~\cite{kerbl3Dgaussians} model employs anisotropic 3D Gaussian ellipsoids as an explicit 3D scene representation, achieving the strongest performance across various tasks in static scenes. StreetGaussian\cite{Yan2024StreetGF}, the latest Gaussian-based method, introduces time into SH coefficients, reaching SOTA performance as well, albeit also utilizing 2D tracked boxes. 
For a fair comparison, we also apply LiDAR point cloud initialization to 3DGS, and depth regularization to 3DGS and MARS, mirroring our approach. 

\textbf{Implementation Details.}
We train our model for 50,000 iterations using the Adam optimizer~\cite{kingma2017adam}, following the learning rate configurations of 3D Gaussians~\cite{kerbl3Dgaussians}. Additionally, we employ 5,000 steps of pure static 3D Gaussian training~\cite{kerbl3Dgaussians} as a warm-up for the scene~\cite{Wu20234DGS}, as illustrated in Figure~\ref{fig:warmup}. For the reconstruction of long sequence scenes, we divide the scene into multiple clips. Specifically, we use 50 frames per clip, where the optimized Spatial-temporal field serves as the initialization for the Spatial-temporal field of the next sequence with 50 steps. This approach maintains spatial and temporal consistency across sequences within the same scene. 
The basic resolution for our multi-resolution HexPlane encoder is set to 64, then upsampled by 2 and 4 as~\cite{Wu20234DGS}. The learning rate of it is set as $1.6 \times 10^{-3}$, decayed to $1.6 \times 10^{-4}$ at the end of trainging. Each decoder in the multi-head decoder is a small MLP with the same learning rate as the HexPlane encoder. Other hyperparameters are kept consistent with 3DGS\cite{kerbl3Dgaussians}.
In the experiments conducted on the Waymo-NOTR dataset, we strictly adhered to the experimental settings of EmerNeRF~\cite{yang2023emernerf}. Similarly, for the Waymo-Street dataset, our experimental setup closely followed StreetGaussian~\cite{Yan2024StreetGF}.

\begin{table}[t]
\centering
\caption{Overall performance of our methods with existing SOTA approaches on the Waymo-NOTR dataset\cite{yang2023emernerf}. 
"PSNR*" and "SSIM*" denote the PSNR and SSIM of dynamic objects respectively.
The \colorbox{lightpink}{best} and the \colorbox{lightblue}{second best} results are denoted by pink and blue.
}
\vspace{0.2em}
\setlength{\tabcolsep}{5pt}
\renewcommand{\arraystretch}{1} 
\begin{tabular}{@{}llcccccccc@{}}
\toprule
\multirow{2}{*}{Data} & \multirow{2}{*}{Metrics} & \multicolumn{4}{c}{Scene Reconstruction} & \multicolumn{4}{c}{Novel View Synthesis} \\ \cmidrule(l){3-6} \cmidrule(l){7-10} 
 & & 3DGS & MARS & EmerNeRF & Ours & 3DGS & MARS & EmerNeRF & Ours \\ \midrule
\multirow{5}{*}{D32} 
 & PSNR$\uparrow$ & \cellcolor{lightblue} 28.47 & 28.24 & 28.16 & \cellcolor{lightpink}31.35
 & 25.14 & \cellcolor{lightblue}26.61 & 25.14 & \cellcolor{lightpink}27.44 \\
 & SSIM$\uparrow$ & \cellcolor{lightblue} 0.876 & 0.866 & 0.806 & \cellcolor{lightpink} 0.911 
 & \cellcolor{lightblue} 0.813 & 0.796 & 0.747 & \cellcolor{lightpink} 0.857 \\
 & LPIPS$\downarrow$ & 0.136 & 0.252 &\cellcolor{lightblue} 0.228 & \cellcolor{lightpink} 0.106
 &\cellcolor{lightblue} 0.165 & 0.305 & 0.313 &\cellcolor{lightpink}0.137 \\
 & PSNR*$\uparrow$ & 23.26 & 23.37 &\cellcolor{lightblue} 24.32 & \cellcolor{lightpink}26.02 
 & 20.48 & 22.21 & \cellcolor{lightpink} 23.49 &\cellcolor{lightblue} 22.92\\
 & SSIM*$\uparrow$ &\cellcolor{lightblue} 0.716 & 0.701 & 0.682 & \cellcolor{lightpink}0.783
 & 0.753 & 0.697 & 0.660 & \cellcolor{lightpink}0.680 \\ \midrule
\multirow{3}{*}{S32}
 & PSNR$\uparrow$ &29.42 & 28.31 & \cellcolor{lightblue}30.00 & \cellcolor{lightpink}30.73& 26.82 & \cellcolor{lightblue}27.63 & \cellcolor{lightpink}28.89 &27.05  \\
 & SSIM$\uparrow$ & \cellcolor{lightpink}0.891 & 0.879 & 0.834 & \cellcolor{lightblue}0.883 & \cellcolor{lightblue}0.836 & \cellcolor{lightpink}0.848 & 0.814 &0.825 \\
 & LPIPS$\downarrow$ & \cellcolor{lightblue}0.118 & 0.196 & 0.201 & \cellcolor{lightpink}0.116 & \cellcolor{lightpink}0.134 & 0.193 & 0.212 & \cellcolor{lightblue}0.142\\
\bottomrule
\end{tabular}
\vspace{-1.8em}

\label{tab:NOTR}
\end{table}

\begin{table}[t]
\centering
\caption{Quantitative results on StreetGaussian datasets~\cite{Yan2024StreetGF}. We strictly follow the experimental setting of it and borrow results from it since it has not been open-sourced.}
\vspace{0.2em}
\setlength{\tabcolsep}{10pt}
\renewcommand{\arraystretch}{1} 
\begin{tabular}{@{}lcccccc@{}}
\toprule
Metrics         & 3D GS &NSG & MARS   & EmerNeRF & StreetGaussian & Ours \\ \midrule
PSNR$\uparrow$  &    29.64   &  28.31  & 31.37  & 32.34 & \cellcolor{lightpink} 34.96& \cellcolor{lightblue}34.61\\
SSIM$\uparrow$  &   0.918    &  0.862  & 0.904  & 0.886 & \cellcolor{lightblue}0.945& \cellcolor{lightpink}0.95z0\\
LPIPS$\downarrow$ &   0.117  &  0.346  & 0.246  & 0.142 & \cellcolor{lightblue}0.068& \cellcolor{lightpink}0.050\\
PSNR*$\uparrow$ &    16.48   &   19.55 & 23.07  & 25.71  &\cellcolor{lightblue}25.46 & \cellcolor{lightpink}25.78\\
\bottomrule
\end{tabular}
 \vspace{-1.6em}

\label{tab:street}
\end{table}

\subsection{Comparisons with the State-of-the-art}
\label{exp: sota}

The results on the Waymo-NOTR dataset demonstrate that our approach consistently outperforms other methods in scene reconstruction and novel view synthesis, as shown in Table~\ref{tab:NOTR}. For the static32 dataset, we utilize PSNR, SSIM, and LPIPS~\cite{Zhang_Isola_Efros_Shechtman_Wang_2018} as metrics to evaluate rendering quality. For the dynamic32 dataset, we additionally include PSNR* and SSIM* metrics focusing on dynamic objects. Specifically, we project the 3D bounding boxes of dynamic objects onto the 2D image plane and calculate pixel loss only within the projected boxes as~\cite{yang2023emernerf, Yan2024StreetGF}. Our metrics outperform those of other existing methods, indicating the superior performance of our approach in modeling dynamic objects. Moreover, although static scene representation is not our primary focus, our method also performs exceptionally well in this aspect. Thus, our approach is more versatile and general.

We also conducted qualitative comparisons, as shown in Figure~\ref{fig:compare}. We emphasized regions with significant differences to provide a clearer demonstration. From the figure, it is evident that our method surpasses the state-of-the-art (SOTA) in both the synthesis of new viewpoints (left side of Figure~\ref{fig:compare}) and reconstruction (right side of Figure~\ref{fig:compare}) of static and dynamic scenes. Although 3DGS~\cite{kerbl3Dgaussians} faithfully reconstructs static objects, it fails when dealing with dynamic objects and struggles with reconstructing distant skies. The reconstruction quality of MARS~\cite{wu2023mars} is poor, being effective only for very short sequences, and it struggles to reconstruct fast-moving objects. While EmerNeRF~\cite{yang2023emernerf} can self-supervise the reconstruction of static and dynamic objects, the reconstruction quality is unsatisfactory, with issues such as ghosting, loss of plant texture details, missing lane markings, and blurry distant scenes. For novel view synthesis, our method can generate high-quality rendered images and ensure consistency between multiple camera views. In dynamic scene reconstruction, we accurately simulate dynamic objects in large-scale scenes, particularly distant dynamic objects, and mitigate issues such as loss, ghosting, or blurriness associated with these dynamic elements.

\begin{figure*}[t]
    \centering
    \includegraphics[width=1\textwidth]{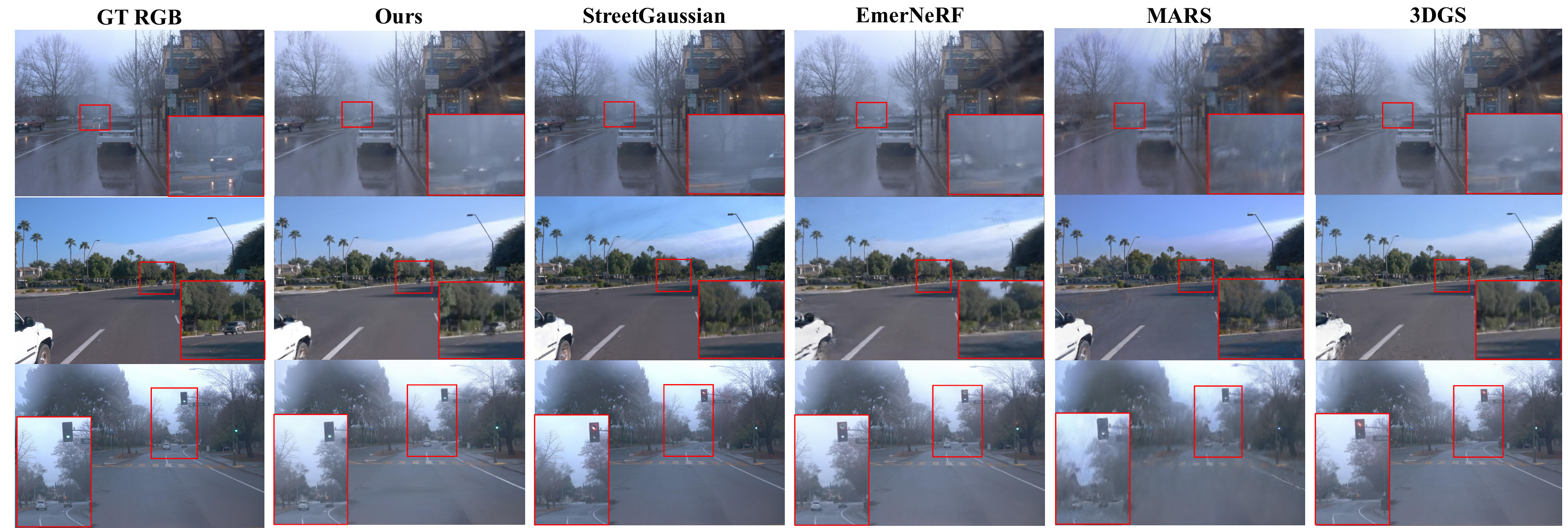}
    \vspace{-1.5em}
    \caption{Qualitative comparison over Waymo-Street Datasets~\cite{Yan2024StreetGF}. All results are from novel view synthesis. Compared to StreetGaussian~\cite{Yan2024StreetGF}, our method demonstrates a stronger ability to self-supervisedly reconstruct distant dynamic objects and is more sensitive to changes in scene details.}
    \label{fig:compare_street}
    \vspace{-1.5em}
\end{figure*}

Table~\ref{tab:street} presents the results on the dataset collected by StreetGaussian~\cite{Yan2024StreetGF}. StreetGaussian is a state-of-the-art method for Gaussian-based dynamic object representation. Our approach performs similarly to StreetGaussian, but with the distinction that StreetGaussian uses additional bounding boxes to model dynamic objects, whereas our approach does not require any explicit supervision. As shown in Figure~\ref{fig:compare_street}, compared to StreetGaussian~\cite{Yan2024StreetGF} which uses explicit supervision, our method excels in self-supervised reconstruction of distant dynamic objects. Additionally, our method is more sensitive to changes in scene details, such as variations in traffic lights. Furthermore, StreetGaussian exhibits noise in the sky, resulting in a decrease in rendering quality.

\begin{table}[t]
\centering
\caption{Quantitative ablation studies on Waymo-NOTR dynamic32 datasets.}
\renewcommand{\arraystretch}{1} 
\begin{tabular}{@{}llccccc|cc@{}}
\toprule
\textbf{Task} & Metrics & \makecell{w/o $\mathcal{P}^{\rho}_{ij}$} & \makecell{w/o $\mathcal{D}_{x}$} & \makecell{w/o $\mathcal{D}_{SH}$} & \makecell{w/o $\mathcal{D}_{s}$} & \makecell{w/o Warm-up} & \makecell{Ours} \\
\midrule
\multirow{4}{*}{\shortstack[l]{Scene\\Reconstruct}} &
PSNR$\uparrow$ &18.702 &29.861 & 31.458 & 31.605 & 31.390& \textbf{32.135}& \\
&SSIM$\uparrow$ & 0.4793& 0.8871& 0.9157& 0.9174& 0.9173 &\textbf{0.9355 }& \\
&PSNR*$\uparrow$ &16.800 &24.626 &26.420 & 26.556 &26.628 & \textbf{27.046}& \\
&SSIM*$\uparrow$ &0.3627 &0.7521 &0.8162 & 0.8182&0.8213 & \textbf{0.8284}& \\
\midrule
\multirow{4}{*}{\shortstack[l]{NVS}} &
PSNR$\uparrow$ & 17.245& 25.850& 27.959& 27.981& 27.955 & \textbf{28.417}& \\
&SSIM$\uparrow$ & 0.4499& 0.8174& 0.8616& 0.8624 &\textbf{0.8641} & \textbf{0.8641}& \\
&PSNR*$\uparrow$ &15.613& 21.385& 21.385& 23.402& 23.681& \textbf{23.974}& \\
&SSIM*$\uparrow$ &0.3118& 0.6386& 0.6386& 0.7138& 0.7117 &\textbf{0.7175} & \\

\bottomrule
\end{tabular}
 \vspace{-1.6em}
\label{tab:ablation}
\end{table}

\subsection{Ablation and Analysis}
\label{exp: ablation}
\begin{figure*}[t]
    \centering
    \includegraphics[width=1\textwidth]{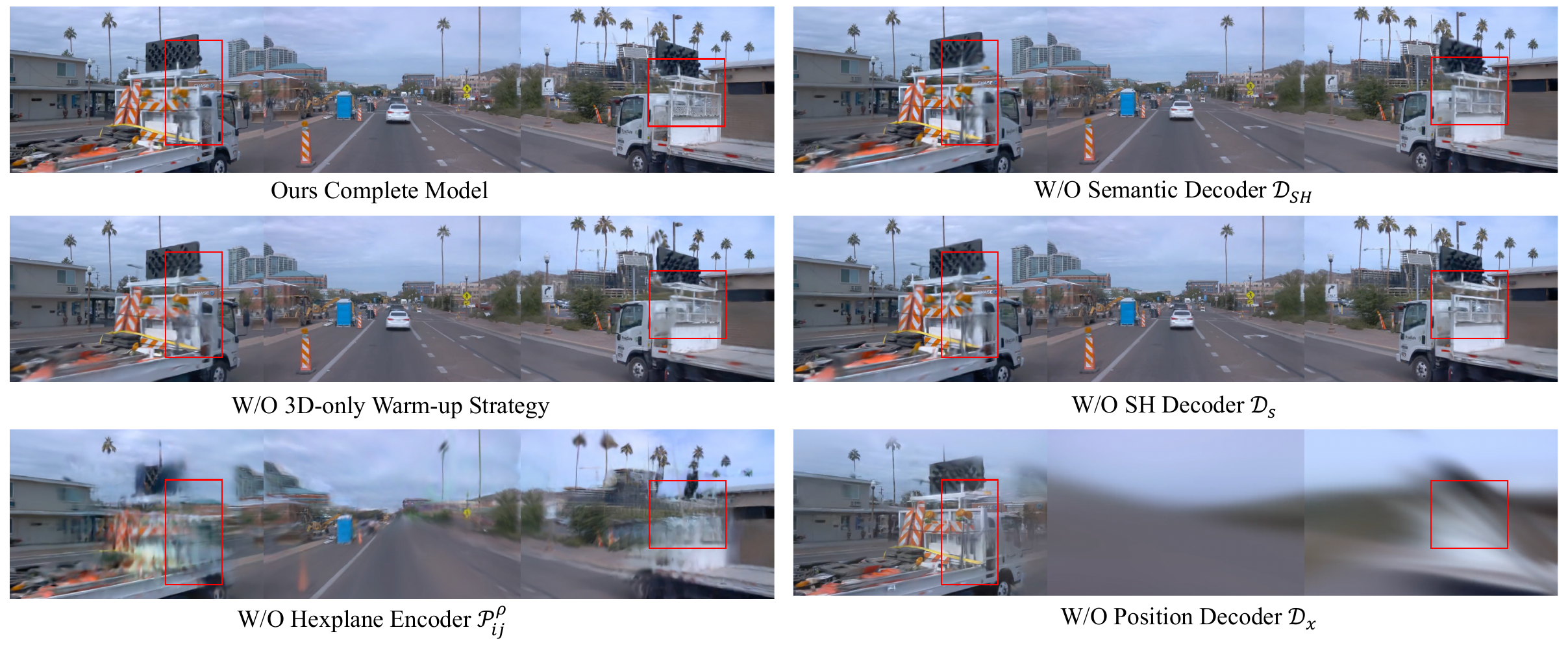}
    \vspace{-1.8em}
    \caption{Visual ablation results on the Waymo-NOTR dynamic32 dataset.}
    \label{fig:ablation}
    \vspace{-1em}
\end{figure*}

 We investigate the effectiveness of our method and its various components. Due to time constraints, we select 20 sequences from NOTR dynamic32~\cite{yang2023emernerf} for analysis, and all models are trained for a shorter duration of $30,000$ iterations. Table~\ref{tab:ablation} presents the quantitative results, while Figure~\ref{fig:ablation} showcases the visual comparison results.
 
\begin{figure*} [t]
    \centering
    \includegraphics[width=1\textwidth]{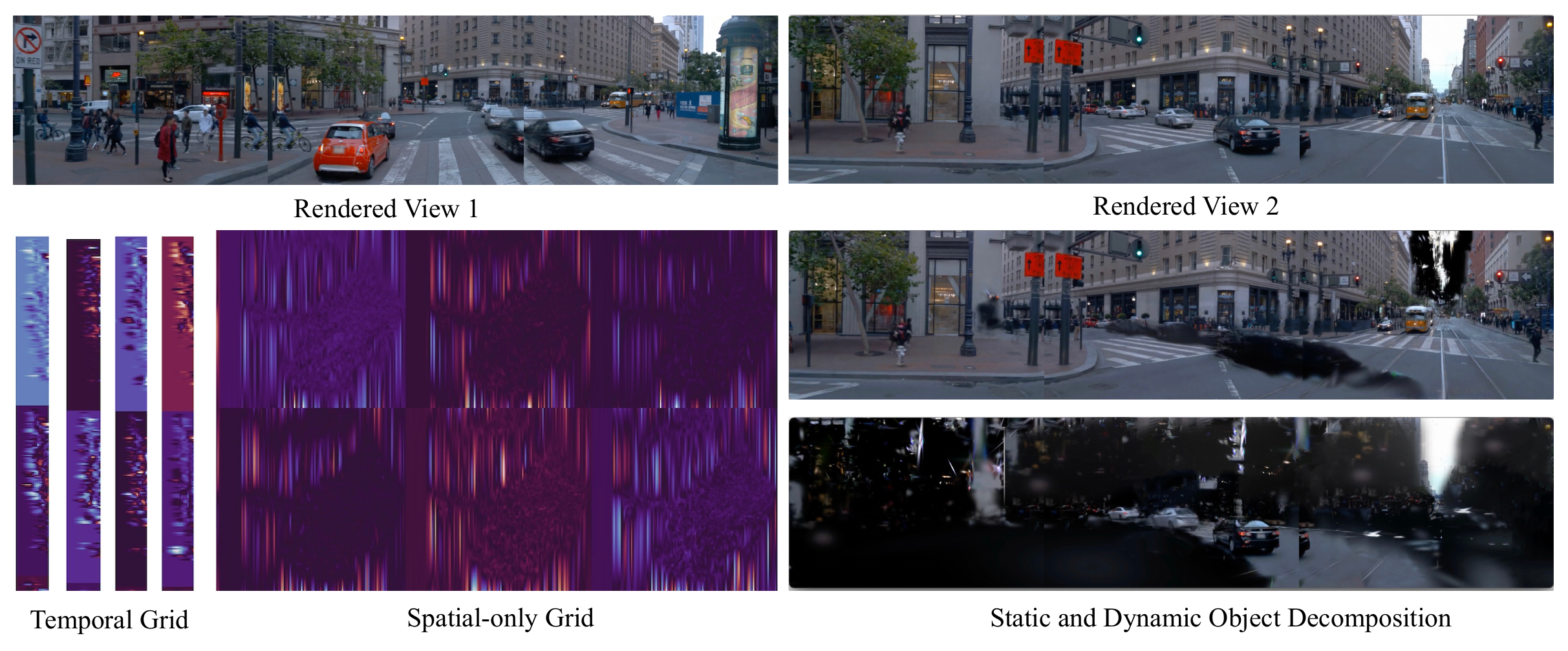}
    \vspace{-2.0em}
    \caption{Visualization of HexPlane voxel grids, showcasing its capability to decompose static and dynamic elements. Spatial-only grid refers to the spatial voxel parameters, while the temporal grid refers to its time features.}
    \label{fig:hexplane}
    \vspace{-2em}
\end{figure*}

\textbf{Multi-resolution Hexplane Structure Encoder.} 
Compared to purely explicit methods, the proposed HexPlane encoder $\mathcal{P}^{\rho}_{ij}$ allows for memory savings and enables retention of different dimensions of spatial-temporal information in the scene through various resolutions. Discarding this module and relying solely on a shallow MLP $\phi _m$ fails to accurately establish spatial-temporal fields and cannot simulate Gaussian deformations. Both Table~\ref{tab:ablation} and Figure~\ref{fig:ablation} demonstrate this, without this module, our rendering quality sharply declines.
We also provide visualizations of the features of this encoder, as shown in Figure~\ref{fig:hexplane}. As an explicit module, we can easily optimize all Gaussian features on a single voxel plane. 
From Figure~\ref{fig:hexplane}, it is evident that the voxel plane features mainly concentrate on the moving parts of the scene. 
The trajectories of moving vehicles in the scene extend from the bottom-right to the top-right corner. As a result, spatial plane features are primarily concentrated in the bottom-right corner, whereas temporal plane features are predominantly observed on the right side. These patterns demonstrate that our encoder successfully captures both spatial and temporal information. This capability allows us to effectively self-supervise the decomposition of static and dynamic components, as illustrated in Figure~\ref{fig:hexplane} and Figure~\ref{fig:pipeline}.

\textbf{Multi-head Gaussian Decoder.} 
Our proposed multi-head Gaussian decoder can decode voxel features. As indicated in Table~\ref{tab:ablation}, disabling this component would impact rendering quality greatly. Additionally, as shown in Figure~\ref{fig:ablation}, disabling the $\mathcal{D}_{x}$ decoder and only training Gaussian in canonical space would introduce significant noise.
The noise stems from Gaussian points initialized by LiDAR point clouds, resulting in a series of Gaussian points along a moving vehicle's trajectory. If these points are not deformed, it becomes challenging to optimize them afterward.
Furthermore, omitting the semantic feature decoder $\mathcal{D}_{s}$ and color deformation decoder $\mathcal{D}_{SH}$ primarily affects rendering details. For example, the geometric structure of the truck becomes blurrier without these components.

\textbf{Static Gaussian Warm-up.} 
According to Figure~\ref{fig:ablation}, we found that directly training the 4D Gaussians without first optimizing 3D Gaussians for warm-up not only reduces convergence speed but also affects the final rendering quality. As shown in the~\ref{fig:warmup}, performing a warm-up step already yields basic static scene reconstruction, which alleviates the pressure on the 4D spatial-temporal network to learn large-scale scenes and allows the network to focus more on dynamic parts. Additionally, it stabilizes the network by avoiding early-stage numerical errors~\cite{Wu20234DGS}.

\section{Conclusion}
In this paper, we propose $S^3$Gaussian, the first self-supervised street Gaussian method to differentiate dynamic and static elements in complex driving scenes. $S^3$Gaussian employs a Spatial-temporal Field Network to achieve the scene decomposition automatically, which consists of a Multi-resolution Hexplane Structure Encoder and a Multi-head Gaussian Decoder. Given a 4D grid in global space, the proposed Hexplane encoder aggregates features into dynamic or static planes. Then we decode these features into the deformed 4D Gaussians. The entire pipeline is optimized without any extra annotations. Experiments on challenging datasets including NOTR and Waymo improve that $S^3$Gaussian show superior scene decomposition ability and obtain the state-of-the-art rendering quality across different tasks. Abundant quantitative results are implemented to shed light on the effectiveness of each component in $S^3$Gaussian.

\newpage

\appendix

\section{Appendix}

\subsection{Additional Implementation Details}
\textbf{Datasets Details.}
Our Waymo-NOTR dataset follows the setup of \cite{yang2023emernerf}. For camera images, we utilize three frontal cameras: FRONT LEFT, FRONT, and FRONT RIGHT, adjusted to a resolution of 640 $\times$ 960 for training and evaluation. The length of all sequences is set to 100 frames. We select every 10th frame from the sequences as the test frames and use the remaining frames for training.
For our Waymo-Street dataset, consistent with~\cite{Yan2024StreetGF}, we use frontal cameras and downscale the input images to 1066 $\times$ 1600 for evaluating monocular reconstruction and novel view synthesis capabilities. The length of all sequences strictly follows the dataset setting released by StreetGaussian~\cite{Yan2024StreetGF}, with each sequence approximately 100 frames long. We select every 4th frame from the sequences as the test frames and use the remaining frames for training.

\textbf{Feature Extraction.} 
We employ the DINOv2~\cite{oquab2023dinov2} checkpoint and the feature extractor implementation by~\cite{amir2021deep}. Specifically, we use the ViT-B/14 variant and adjust the image dimensions to 644$\times$966 with a stride of 7. Given the large size of the feature maps, following~\cite{yang2023emernerf} we use PCA decomposition to reduce the feature dimension from 768 to 3 and normalize these features to the [0,1] range.

\subsection{More Related Work (Advances in Neural Radiance Fields)}
In recent years, there has been a surge of interest among researchers in leveraging neural rendering techniques for scene modeling. Among these techniques, Neural Radiance Fields (NeRF) have garnered particular attention. NeRF~\cite{mildenhall2021nerf} utilizes differentiable volume rendering methods, facilitating the generation of novel scenes from a mere collection of planar images accompanied by their respective camera poses. Moreover, NeRF demonstrates the capability to segregate street views into static and dynamic scenes by tracking the bounding boxes of vehicles. Despite the extensive research efforts aimed at enhancing NeRF's functionalities, which have led to notable advancements in training speed~\cite{Fridovich-Keil_Meanti_Warburg_Recht_Kanazawa_2023, Garbin2021FastNeRFHN, Muller_Evans_Schied_Keller_2022}, pose optimization~\cite{Bian2022NoPeNeRFON, Lin2021BARFBN,Wang2021NeRFNR}, scene editing~\cite{Li2022ClimateNeRFEW, Rudnev2021NeRFFO}, object generation~\cite{huang2024customizeit3d}, and dynamic scene representation~\cite{Huang2021HDRNeRFHD, Pumarola2020DNeRFNR}, challenges persist, particularly regarding training and rendering speed. These challenges pose significant obstacles to the widespread adoption of NeRF in autonomous driving scenarios. Compared to NeRF-based methods, $S^3$Gaussians proposed 4D Gaussian representations for dynamic scenes, significantly boosting rendering speed.

\subsection{Limitations}
Similar to other methods~\cite{yang2023emernerf,wu2023mars}, our scene encounters difficulty in modeling objects moving at high speeds. We suspect this may be due to the deformation field's high variance, rendering it unable to model their rapid movements accurately. Moreover, views of rapidly moving dynamic objects are typically sparse, with only a few views available for capture, making reconstruction even more challenging. 
How to reconstruct these challenging scenes will be a focus of our future research.

\bibliographystyle{ieeenat_fullname} 
\bibliography{ref}

\end{document}